\documentclass[a4paper,conference]{IEEEtran}
\usepackage[breaklinks=true,bookmarks=false]{hyperref}
\usepackage{amsmath}
\usepackage{amssymb}
\usepackage{graphicx}
\usepackage{subfigure}
\usepackage{cite}
\usepackage{booktabs}

\ifCLASSINFOpdf
\else
\fi
\begin{document}
\IEEEoverridecommandlockouts
%
\title{Gaussian Constrained Attention Network for Scene Text Recognition}


\author{\IEEEauthorblockN{Zhi Qiao\textsuperscript{1}$^,$\textsuperscript{2},
Xugong Qin\textsuperscript{1}$^,$\textsuperscript{2},
Yu Zhou\textsuperscript{1}$^,$\IEEEauthorrefmark{1},
Fei Yang\textsuperscript{3},
Weiping Wang\textsuperscript{1}}
\IEEEauthorblockA{\textsuperscript{1}Institute of Information Engineering, Chinese Academy of Sciences}
\IEEEauthorblockA{\textsuperscript{2}School of Cyber Security, University of Chinese Academy of Sciences}
\IEEEauthorblockA{\textsuperscript{3}TAL Education Group}
Email: \{qiaozhi, qinxugong, zhouyu, wangweiping\}@iie.ac.cn, yang.fei@100tal.com\\
\thanks{\IEEEauthorrefmark{1}The corresponding author}}



%


\maketitle

\begin{abstract}
Scene text recognition has been a hot topic in computer vision. Recent methods adopt the attention mechanism for sequence prediction which achieve convincing results. However, we argue that the existing attention mechanism faces the problem of attention diffusion, in which the model may not focus on a certain character area. In this paper, we propose Gaussian Constrained Attention Network to deal with this problem. It is a 2D attention-based method integrated with a novel Gaussian Constrained Refinement Module, which predicts an additional Gaussian mask to refine the attention weights. Different from adopting an additional supervision on the attention weights simply, our proposed method introduces an explicit refinement. In this way, the attention weights will be more concentrated and the attention-based recognition network achieves better performance. The proposed Gaussian Constrained Refinement Module is flexible and can be applied to existing attention-based methods directly. The experiments on several benchmark datasets demonstrate the effectiveness of our proposed method. Our code has been available at https://github.com/Pay20Y/GCAN.
\end{abstract}


%
\IEEEpeerreviewmaketitle

\section{Introduction}
\label{ref_intro}

Nowadays, scene text detection and recognition have attracted more and more research interests. Text detection~\cite{liu2014text,tian2016detecting,zhou2017east,qin2019curved,chen2019constrained} is the task of localizing the text instances in the scene images. Text recognition aims to transcribe the word in the image to computer editable text, which plays an important role, and many methods~\cite{bissacco2013photoocr,shi2013scene,gomez2014scene,jaderberg2014synthetic,rong2014scene,zhou2014perspective,mishra2016enhancing,Shi2016An,wang2017gated,yang2017adadnns,wang2017sequence,busta2017deep,shi2018aster,qiao2020seed,sun2018textnet,gao2018dense,bartz2018see,zhang2018radical,wan20192d,luo2019moran,li2019show,baek2019wrong,feng2019textdragon,zhang2019sequence,wei2019new,cong2019comparative,kong2019garn,nag2019crnn,chen2019modified,zheng2019mining} achieve convincing results. Most of recent methods adopt the attention mechanism, which let the model learn the character alignments in a weak-supervised way. The attention mechanism is usually integrated into the encoder-decoder framework, where the encoder is to extract rich visual features and the decoder is to transcribe the sequence step by step. Specifically, the attention weights are calculated between the feature map from the encoder and the hidden state of the RNN in the decoder. In this way, the decoder focuses on different local areas to predict the corresponding characters at each time step. According to the different dimensions of the feature map, the existing attention methods can be divided into 1D based and 2D based attention.

\begin{figure}[t]
\begin{center}
   \includegraphics[width=1.0\linewidth]{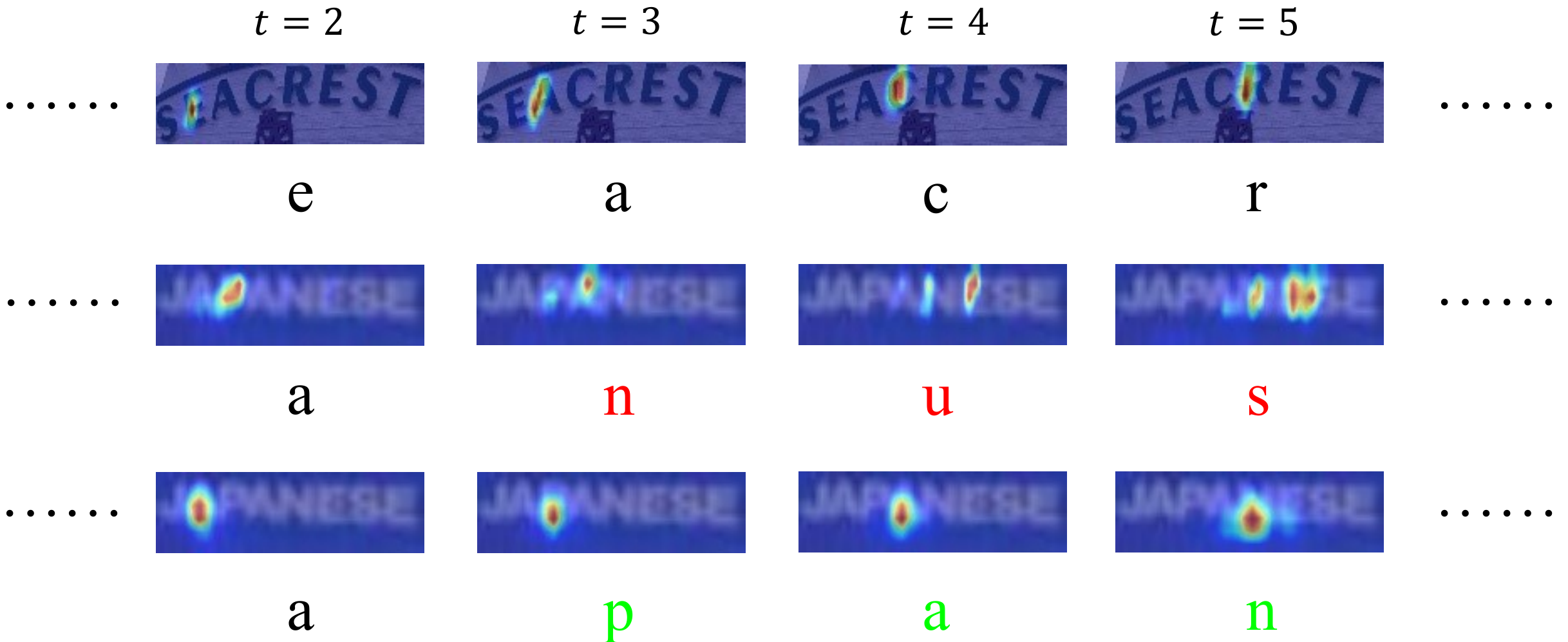}
\end{center}
   \caption{The first row shows the attention maps with a concentrated distribution. The second row shows the attention maps suffered from the problem of \textit{attention diffusion}, in which the noise feature may be introduced into the decoding process and lead to the wrong predictions. The third row shows the attention maps in our proposed method. With the refinement, our proposed method predicts more concentrated attention maps and achieves better performance.}
\label{fig_intro}
\end{figure}

\begin{figure*}[t]
\begin{center}
\includegraphics[width=0.8\linewidth]{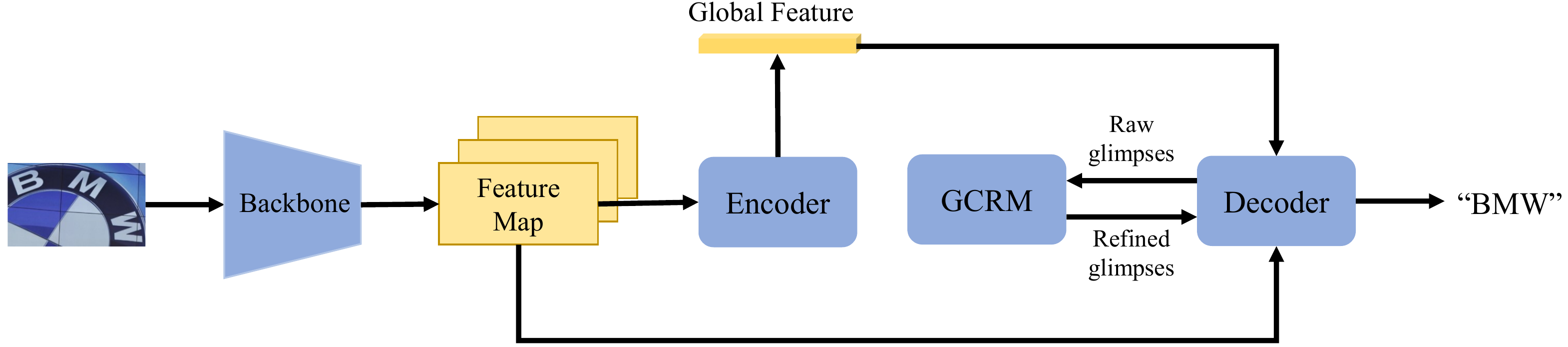}
\end{center}
   \caption{The framework of our proposed GCAN. Different from the traditional 2D attention framework such as~\cite{li2019show}, a proposed GCRM module can refine the raw attention map and generate refined glimpses for better decoding performance.}
   \label{fig_pipeline}
\end{figure*}

The characters in an image are usually with a regular shape, so the attention weights will be a concentrated distribution that seems like the Gaussian distribution as shown in the first row of Fig.~\ref{fig_intro}. It is an obvious characteristic of text recognition different from machine translation and image caption. In the weak-supervised training strategy, the attention weights are calculated from the whole feature map at every time step without explicit constraints, where this characteristic is not fully used. As shown in the second row in Fig.~\ref{fig_intro}, the attention weights are not in a concentrated distribution, and the model will suffer a lot from the diffuse alignment. In this paper, we call this problem \textit{attention diffusion}, where the noise feature will be introduced into the decoding process.

There are several works~\cite{yang2017learning,cheng2017focusing,he2018anend-to-end} focusing on the quality of the attention weights. Specifically, \cite{yang2017learning} generates Gaussian distribution labels for each characters to supervise the attention weights. \cite{cheng2017focusing} proposes focusing attention network to solve the problem of attention drift. \cite{he2018anend-to-end} introduces two additional losses to the center of the character and the character mask for better attention alignment. However these works lack an explicit process to constraint the attention, and they may still generate inaccurate attention weights in some situations.

In this paper, we introduce a new module named Gaussian Constrained Refinement Module (GCRM) to the existing 2D attention mechanism. Integrated with 2D attention based method SAR~\cite{li2019show}, we propose our new method Gaussian Constrained Attention Network (GCAN) as shown in Fig.~\ref{fig_pipeline}. Specifically, the raw attention weights are calculated between the hidden states of the RNN in the decoder and the feature map generated by the encoder. The glimpses are generated with the weighted sum between attention weights and the feature map. Different from the traditional attention mechanism, our proposed method predicts the additional Gaussian parameters (i.e. the mean and variance) from the hidden state and corresponding glimpse, then the Gaussian masks are constructed by the predicted parameters. To refine the raw attention weights, the Gaussian masks are applied to them with element-wise multiplication. Finally, the refined attention weights are used to generate the new glimpses with more accurate alignment, then we fuse the two glimpses to predict the corresponding characters at each time step. The process is illustrated in Fig.~\ref{fig_decoder} and Fig.~\ref{fig_gcrm}. Notice that, the proposed GCRM is flexible and easy to be integrated into any attention-based framework. As shown in the third row in Fig.~\ref{fig_intro}, the attention weights are more concentrated and lead to better decoding performance with the refinement.

The main contributions are summarized as follows:

1) We define and explore the problem of \textit{attention diffusion} in scene text recognition, where the attention weights are not concentratedly distributed and will bring noise features into the decoding process.

2) Different from only additional supervision in the training process, we propose the GCRM, which explicitly predicts the Gaussian mask to refine the attention weights. The method is explicit, flexible and easy to implement.

3) The proposed GCRM improves the performance of the traditional attention mechanism. Integrating SAR~\cite{li2019show} with GCRM, our proposed GCAN achieves state-of-the-art results on several scene text recognition benchmarks.

\section{Related Work}
\label{ref_relate}

\subsection{Scene Text Recognition}
Scene text recognition has been studied for many years, and the existing methods can be divided into traditional methods and deep learning based methods. 

For traditional methods~\cite{wang2010word,wang2011end,mishra2012top,novikova2012large,yao2014strokelets,bai2016strokelets}, they usually adopt a bottom-up framework, which detects and classifies single character first, then groups them with heuristic rules, lexicons or language model. 

Nowadays, with the development of CNN and RNN, deep learning based methods~\cite{Jaderberg2016Reading,he2016reading,Shi2016An,su2017accurate,yin2017scene,lee2016recursive,shi2016robust,yang2017learning,cheng2017focusing,wu2018scan,cheng2018aon,shi2018aster,li2019show,liao2019scene,luo2019moran,yang2019symmetry} have been the mainstream methods. \cite{Jaderberg2016Reading} regards the text recognition as a classification task with the CNN. It is effective but not flexible because of the pre-defined vocabulary. To overcome this limitation, many recent methods treat text recognition as a sequence prediction task. They can be divided into two main approaches according to decoding manners, Connectionist Temporal Classiﬁcation (CTC) based and attention mechanism based. For CTC based methods,~\cite{he2016reading,Shi2016An,su2017accurate} adopt RNN for sequence modeling and CTC for feature alignment. They are effective and fast with a highly integrated network.~\cite{yin2017scene} proposes to detect and classify characters with sliding windows, then uses CTC to decode the sequence. Recently, inspired by machine translation and image caption, attention mechanism is introduced into text recognition. \cite{lee2016recursive} proposes a recursive CNN network to capture broader features and an attention-based decoder used to transcribe sequence. \cite{cheng2017focusing} introduces a focusing attention network to deal with the problem of \textit{attention drift}, and gets more accurate alignments. \cite{he2018anend-to-end} adds supervision on the center of characters and characters mask to enhance the attention alignment.

Based on convincing progress on horizontal text, many works try to recognize the text of irregular shapes. They can be mainly divided into four categories, rectification based, 2D attention based, segmentation based and multi-direction encoding based. Rectification based methods adopt an intuitive way which rectifies irregular shape of text to horizontal. \cite{shi2016robust,shi2018aster} predict the control points in a weak-supervised way, and adopt spatial transform network~\cite{jaderberg2015spatial} to rectify the text. \cite{zhan2019esir} achieves better performance with iterative rectification. \cite{yang2019symmetry} predicts some constrained geometry attributes instead of predicting control points directly and generates better rectification results. \cite{luo2019moran} adopts offset map to rectify irregular text. Instead of rectifying the whole text, \cite{liu2018char} detects and rectifies each character. The 2D attention based methods deal with the irregular text directly with 2D attention mechanism. \cite{yang2017learning} first introduces 2D attention to text recognition, and adopts character-level supervision to guide the attention. \cite{li2019show} proposes tailored 2D attention operation without pixel-level or character-level annotations. \cite{wang2020decouple} proposes a convolutional alignment module to predict attention weights parallel. \cite{liao2019scene} is a segmentation-based method, which treats the recognition as semantic segmentation with fully convolution network~\cite{long2015fully}. For multi-direction encoding, \cite{cheng2018aon} encodes four direction features with two LSTM and a filter gate is proposed to fuse the four direction features. Apart from the four main technologies, \cite{xie2019aggregation} proposes a new loss function ignoring the sequence order which is effective with fast decoding speed.

\subsection{Quality of Attention}
As mentioned before, \cite{yang2017learning,cheng2017focusing,he2018anend-to-end} focus on the quality of the attention weights. \cite{yang2017learning} generates Gaussian distribution labels for each characters to supervise the attention weights. \cite{cheng2017focusing} computes the center points of the attention and focuses the attention distribution into the target localization. \cite{he2018anend-to-end} supervises the characters center and the characters mask. Besides text recognition, \cite{Min2019Attention} proposes Attention Refine Unit to retain the attention on the salient parts and restrain the attention on irrelevant parts. \cite{jaeyoung2019tgsa} and \cite{duan2019attention} adopt attention mask for the self-attention in the encoder to extract richer local features. \cite{luong2015effective} proposes the local attention to let the model focus on a window of its neighbors. The local attention lacks flexibility because of the fixed size of window, and it is not suitable for text recognition. Because the scale of characters in images are various, it is difficult to select the proper size of window.

Inspired by these works, we propose to predict a Gaussian mask to refine the attention weights explicitly. In the next section, we will describe the pipeline of our proposed GCAN and the details of the GCRM. 


\section{Methods}
\label{ref_method}
In this section, we introduce our proposed GCAN in detail. In \ref{ref_pipeline} we introduce the pipeline of the method. In \ref{ref_decoder} we show the framework of the decoder and in \ref{ref_gcrm} we describe the details of our proposed GCRM which is to refine the raw attention weights from the decoder. 

\subsection{Pipeline}
\label{ref_pipeline}

As shown in Fig.~\ref{fig_pipeline}, our proposed GCAN contains four major components. the backbone adopts a 31-layer ResNet designed by~\cite{li2019show}, which is suitable for 2D attention mechanism. The encoder encodes a global feature to guide the decoding process. Same as~\cite{li2019show}, we apply vertical global pooling to the feature map generated from the backbone, then a 2-layer LSTM is to capture the context information. The last hidden state of the LSTM is used as the global feature, it will be fed to the first time step of the decoding process. These two modules are same as \cite{li2019show}, so we will not describe them in detail. Decoder is another 2-layer LSTM equipped with Luong Attention~\cite{luong2015effective}, and GCRM predicts a Gaussian mask to refine the raw attention weights from the decoder at each decoding step. These two modules will be illustrated in the next two sub-sections respectively. 

\subsection{Decoder}
\label{ref_decoder}

The decoder adopts 2-layer LSTM with the 2D attention mechanism. At each time step, the character predicted by the previous time step is input to LSTM after an embedding layer, and we adopt teacher forcing strategy in the training phase where we replace the predicted character with the corresponding one from ground-truth label. After feeding the embedding of the character, the hidden state of the LSTM will be updated, and it will work as a query to participate in the attention weights calculation as follows,

\begin{equation}
    \alpha_{t}=softmax(W_a^Ttanh(W_{h}h_t + W_{f}f) + b_a)\,.
    \label{eq_att}
\end{equation}
where, $ W_a, b_a, W_h, W_f $ are the trainable weights, $ \alpha_{t} $ is the attention weight at the time step {t}, $ h_t $ is the hidden state of the LSTM, $ f $ is the feature map generated from the backbone. With the attention weights $ \alpha $, the glimpses are calculated with the element-wise multiplication and the sum between the attention weights and the feature map as follows,

\begin{figure}[t]
\begin{center}
\includegraphics[width=0.8\linewidth]{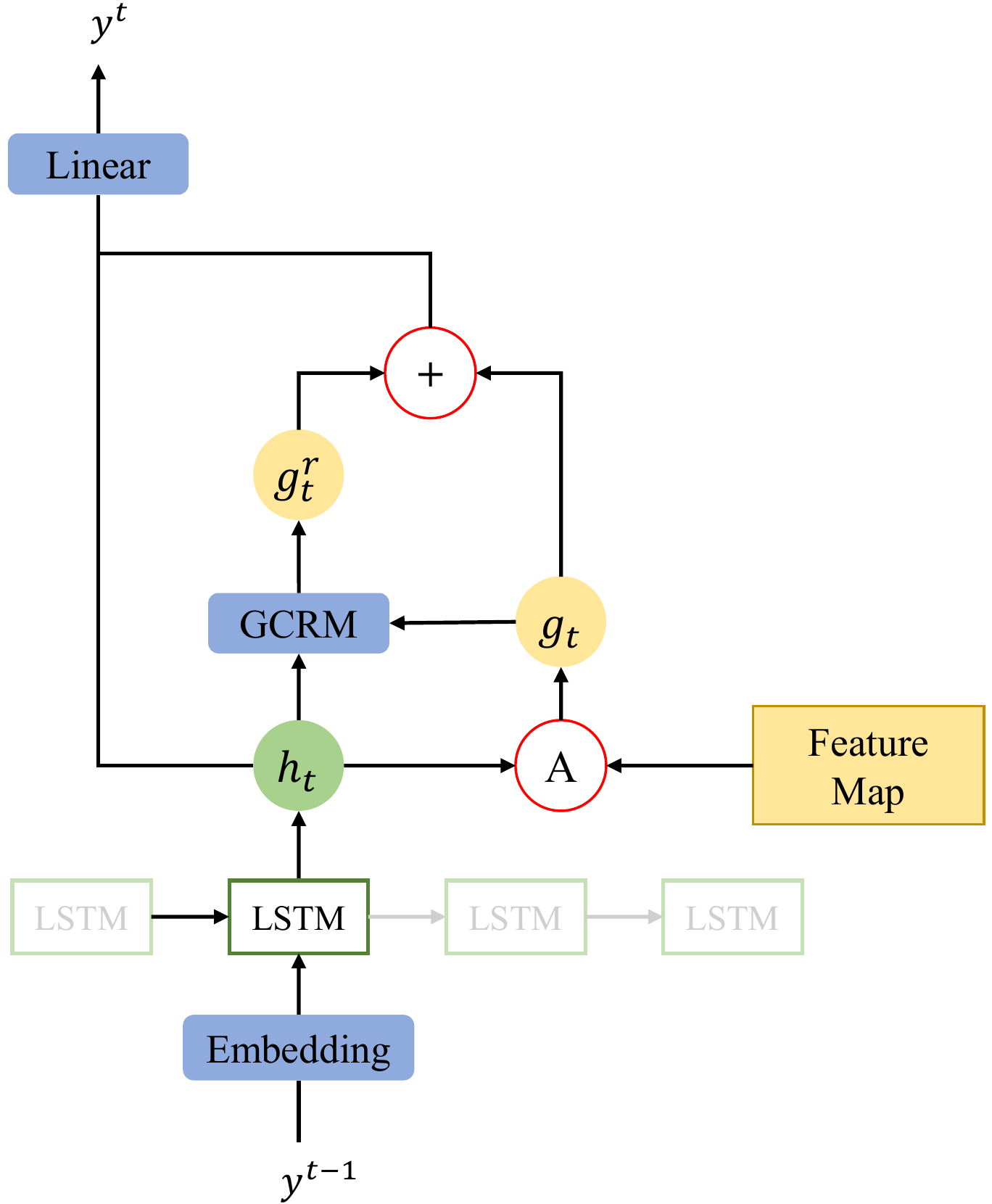}
\end{center}
   \caption{The process of the decoder integrated with GCRM at the $t$ decoding step. $A$ represents the attention operation illustrated in Eq.~\ref{eq_att} and Eq.~\ref{eq_glim}.}
   \label{fig_decoder}
\end{figure}

\begin{equation}
    g_t=\sum_{i,j}\alpha_t^{ij}f_{ij}\,.
    \label{eq_glim}
\end{equation}
where $ g_t $ is the glimpse, $i$ and $j$ is the coordinates along width and height. 
For the conventional attention mechanism, the glimpses will be used to predict the corresponding character directly. However, as we discussed before, the attention weights are calculated from the whole feature map without any constraint that may suffer the problem of \textit{attention diffusion}. So we propose a new module to refine the raw attention weights and generate corresponding refined glimpses $ g_t^r $. The details are described in the next sub-section. Finally, we fuse the two glimpses to make the final prediction:

\begin{equation}
    y^t=softmax(W_y[h_t;(g_t+g_t^r)]+b_y)\,.
\end{equation}
where $ y^t $ is the probability of the output characters, $ W_y $ and $ b_y $ are the trainable parameters, and $ [;] $ is the concatenate operation. The whole process of one single decoding step is illustrated in Fig.~\ref{fig_decoder}.



\begin{figure}[t]
\begin{center}
\includegraphics[width=0.8\linewidth]{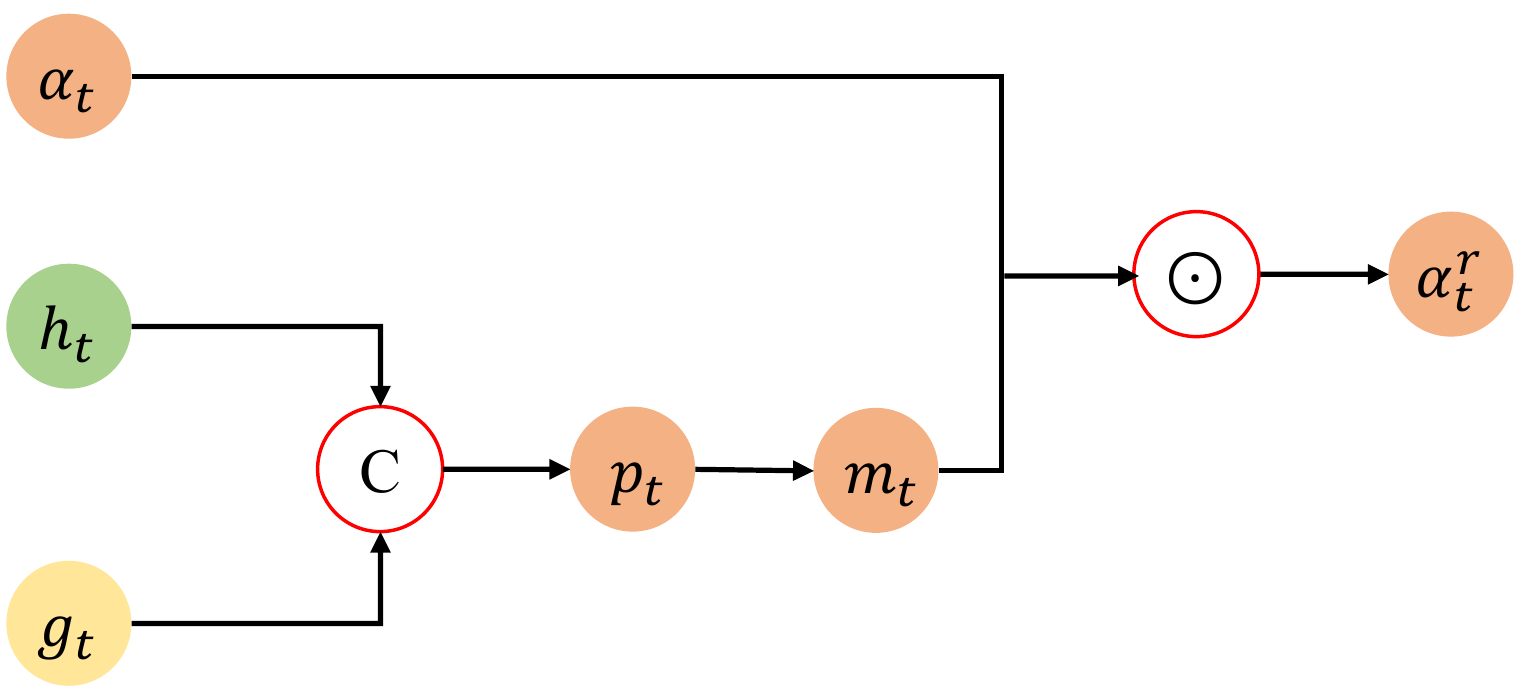}
\end{center}
   \caption{The illustration of the refinement at time step $t$. $C$ represents the concatenate operation and $m_t$ represents the Gaussian mask $mask_t$.}
   \label{fig_gcrm}
\end{figure}

\subsection{Gaussian Constrained Refinement Module (GCRM)}
\label{ref_gcrm}

In this sub-section, we introduce the GCRM in detail. As shown in Fig.~\ref{fig_gcrm}, the GCRM predicts the Gaussian mask with the input of $ h_t $ and $ g_t $ to refine the attention weight, then the refined glimpse are generated accordingly. The parameters of the Gaussian mask are predicted by the $ h_t $ and the $ g_t $: 

\begin{equation}
    \begin{split}
    &p_t=sigmoid(W_p[h_t;g_t]+b_p))\,,\\
    &\{\mu^x_t, \mu^y_t, \sigma^x_t, \sigma^y_t\} = p_t\,.
     \end{split}
\end{equation}
$W_p\in\mathbb{R}^{(C_h+C_g)\times4}$ and $b_p$ are the trainable parameters, where the $C_h$ and the $C_g$ indicate the dimensions of the hidden state $h_t$ and the glimpse $g_t$. $p_t$ is the predicted parameters which consist of $\mu^x_t$, $\mu^y_t$, $\sigma^x_t$ and $\sigma^y_t$ indicating the mean and variance along the two directions respectively. Finally, the parameters of the Gaussian distribution are constructed as follows,
\begin{equation}
    \label{eq_param}
    \begin{split}
    \mu^x_t &= w_f\mu^x_t\,,\\
    \mu^y_t &= h_f\mu^y_t\,,\\
    \sigma^x_t &= (w_f^2/4)\sigma^x_t\,,\\
    \sigma^y_t &= (h_f^2/4)\sigma^y_t\,.\\
    \end{split}
\end{equation}
where $w_f$, $h_f$ are the width and height of the feature map.

With the parameters the Gaussian mask can be constructed:
\begin{equation}
    \begin{split}
    &mask^{(i,j)}_t = \mathcal{N}((i,j)^\intercal|M_t,\Sigma_t)\,,\\
    &M_t=(\mu^x_t,\mu^y_t)^\intercal, \Sigma_t=\left[\begin{matrix}
                                                     \sigma^x_t & 0 \\
                                                     0 & \sigma^y_t \\
                                                \end{matrix}\right]\,.
    \end{split}
\end{equation}
where $\mathcal{N}$ represents the Gaussian distribution. $M_t$ and $\Sigma_t$ are the matrices of mean and variance respectively, which are constructed with the predicted parameters at Eq.~\ref{eq_param}. Finally, the Gaussian mask is applied to the raw attention weights:

\begin{equation}
    \begin{split}
    &\alpha_t^r=mask_t \odot \alpha_t\,,\\
    &g_t^r=\sum_{i,j}\alpha_{t,ij}^rf_{ij}\,.
    \end{split}
\end{equation}
$ \odot $ represents element-wise multiplication. The refined glimpses $g_t^r$ are generated with the weighted sum between the refined attention weights and the feature map. Compared with the local attention~\cite{luong2015effective}, the variance of the Gaussian distribution is predicted rather than fixed, which is more flexible and suitable for text recognition.

\subsection{Label Generation and Training Objective}

Note that the process of the refinement can be trained in a weak-supervised way with only the word-level annotations. But here we add additional supervision of the refined attention weights for better performance. Therefore, we construct Gaussian distribution label for each refined attention weight as follows. Given the bounding box of each character, we first generate the minimum bounding rectangles accordingly, then the center, height, and width are calculated. The mean and variance of the Gaussian distribution are calculated from the center and height (or width) respectively similar as~\cite{yang2017learning}. 

Our model is end-to-end training with the objective function:

\begin{equation}
    L = L_{recog} + \lambda L_{att}\,.
\end{equation}
where $L_{recog}$ is the standard cross-entropy loss of the predicted probabilities with respect to the label of transcription, and $L_{att}$ is the smooth-L1 loss between the refined attention weights $\alpha^r$ and the Gaussian distribution labels. $\lambda$ is used to balance the loss, and we set $\lambda=10$ in our implementation. For datasets without character-level annotations, the attention weights loss $L_{att}$ is ignored simply.


\section{Experiments}
\label{experiment}

In this section, we conduct extensive experiments to verify the effectiveness of our proposed method. In \ref{dataset} we introduce the datasets we used, and we show the details of the implementation in \ref{implementation}. We compare several previous works in \ref{performance}, and we also perform ablation study in \ref{ablation} to show the effectiveness of our proposed GCRM. The speed of training and inference are analyzed in~\ref{speed}. Finally, we analyse the performance of different text length in~\ref{length}. 

\begin{table*}[t]
    \caption{Lexicon-free performance on public benchmarks. \textbf{Bold} represents the best performance. \underline{Underline} represents the second best result.}
    \label{tabel_sota}
    \centering
    \begin{tabular}{|l|c|c|c|c|c|c|}
    \hline 
    Methods
    & IIIT5K & SVT & IC13 & IC15 & SVTP & CUTE \\ 
    \hline
    \hline
    Shi \textit{et al.}~\cite{Shi2016An} & 81.2 & 82.7 & 89.6 & - & - & - \\
    Shi \textit{et al.}~\cite{shi2016robust} & 81.9 & 81.9 & 88.6 & - & 71.8 & 59.2 \\
    Lee \textit{et al.}~\cite{lee2016recursive} & 78.4 & 80.7 & 90.0 & - & - & - \\
    Yang \textit{et al.}~\cite{yang2017learning} & - & - & - & - & 75.8 & 69.3 \\
    Cheng \textit{et al.}~\cite{cheng2017focusing} & 87.4 & 85.9 & 93.3 & 70.6 & - & - \\
    Cheng \textit{et al.}~\cite{cheng2018aon} & 87.0 & 82.8 & - & 68.2 & 73.0 & 76.8 \\
    Liu \textit{et al.}~\cite{liu2018char} & 92.0 & 85.5 & 91.1 & 74.2 & 78.9 & - \\
    Bai \textit{et al.}~\cite{bai2018edit} & 88.3 & 87.5 & \textbf{94.4} & 73.9 & - & - \\
    Liu \textit{et al.}~\cite{liu2018squeezedtext} & 87.0 & - & 92.9 & - & - & - \\
    Liu \textit{et al.}~\cite{liu2018synthetically} & 89.4 & 87.1 & \underline{94.0} & - & 73.9 & 62.5 \\
    Shi \textit{et al.}~\cite{shi2018aster} & 93.4 & 89.5 & 91.8 & 76.1 & 78.5 & 79.5 \\
    Liao \textit{et al.}~\cite{liao2019scene} & 91.9 & 86.4 & 91.5 & - & - & 79.9 \\
    Zhan \textit{et al.}~\cite{zhan2019esir} & 93.3 & \textbf{90.2} & 91.3 & 76.9 & 79.6 & 83.3 \\
    Xie \textit{et al.}~\cite{xie2019aggregation} & - & - & - & 68.9 & 70.1 & 82.6 \\
    Li \textit{et al.}~\cite{li2019show} & 91.5 & 84.5 & 91.0 & 69.2 & 76.4 & 83.3 \\
    Luo \textit{et al.}~\cite{luo2019moran} & 91.2 & 88.3 & 92.4 & 74.7 & 76.1 & 77.4 \\
    Yang \textit{et al.}~\cite{yang2019symmetry} & \textbf{94.4} & 88.9 & 93.9 & \textbf{78.7} & \underline{80.8} & \textbf{87.5} \\ 
    Wang \textit{et al.}~\cite{wang2020decouple} & 94.3 & 89.2 & 93.9 & 74.5 & 80.0 & 84.4 \\
    \hline
    SAR-reproduced & 93.0 & 86.7 & 90.8 & 76.1 & 77.0 & 83.7 \\ 
    GCAN (Ours) & \textbf{94.4} & \underline{90.1} & 93.3 & \underline{77.1} & \textbf{81.2} & \underline{85.6} \\
    \hline
    \end{tabular}
\end{table*}

\subsection{Datasets}
\label{dataset}

\textbf{IIIT5K-Words} (IIIT5K)~\cite{mishra2012scene} contains 3000 high-quality images for testing, most of which are horizon. A 50-word and a 1K-word lexicon are provided for each image.

\textbf{Street View Text} (SVT)~\cite{wang2011end} consists of 350 images for the text detection task. There are 100 images for training and 250 images for testing. 647 word images are cropped from the testing images, some of them are suffered from blurry, noises and low-resolution, and each image is associated with a 50-word lexicon.

\textbf{SVT-Perspective} (SVTP)~\cite{quy2013recognizing} is used for evaluating perspective text recognition, which contains 238 images with 645 cropped word images. Most of them are distorted text, which are hard to recognition.

\textbf{ICDAR2013} (IC13)~\cite{karatzas2013icdar} is collected with focusing. It is used for three main tasks, and for text recognition it contains 1015 word images. Most of them are regular shape text with high-quality, no lexicon is provided.

\textbf{ICDAR2015} (IC15)~\cite{karatzas2015icdar} contains 1500 images, where 1000 images for training and 500 images for testing. Because of the capturing without preparation, most of them are blurred and perspective. It is one of the most challenging datasets recently. 

\textbf{CUTE80} (CUTE)~\cite{risnumawan2014robust} is a small dataests with 288 cropped images, which aims to evaluate arbitrary text recognition. Most of them are curved and no lexicon is provided.

\textbf{Synth90K}~\cite{Jaderberg2016Reading} contains 9 million synthetic text images from a 90K word lexicon which includes the words in the testing set of IC13 and SVT. It is widely used for most text recognition methods as a training set, and only word-level annotations are provided. 

\textbf{SynthText}~\cite{gupta2016synthetic} is another synthetic datasets for text detection task. We cropped the word instances with the bounding boxes, and for each word instances word-level and character-level annotations are both provided.

\subsection{Implementation Details}
\label{implementation}

Our proposed method is implemented with TensorFlow~\cite{abadi2015tensorflow:}. 97 symbols are covered for recognition, including digits, case-sensitive letters, 32 punctuation marks, end-of-sequence symbol, padding symbol, and unknown symbol. The size of the input images is $ 48 \times 160 $ with keeping ratio and padding, and the model is trained on Synth90K and SynthText from scratch. Adam~\cite{kingma2014adam} is adopted as the optimizer, and total training iterations is set to $300K$ with batch size of 128. The learning rate is initialized as $10^{-4}$ and decayed to $10^{-5}$ and $10^{-6}$ at the iteration of $180K$ and $240K$ respectively. The model is trained on a single NVIDIA M40 graphics card. 

We use the model trained on two synthetic datasets for evaluation on all six benchmarks directly. Same as the training phase, the input image is resized to $ 48 \times 160 $ with keeping ratio and padding. We select the character of maximum probability at each time step as the predicted result without beam search or any extra correction. Same as the previous works, the evaluation protocol is case insensitive and disregards punctuation.

\begin{figure*}[t]
\centering
\includegraphics[width=0.8\linewidth]{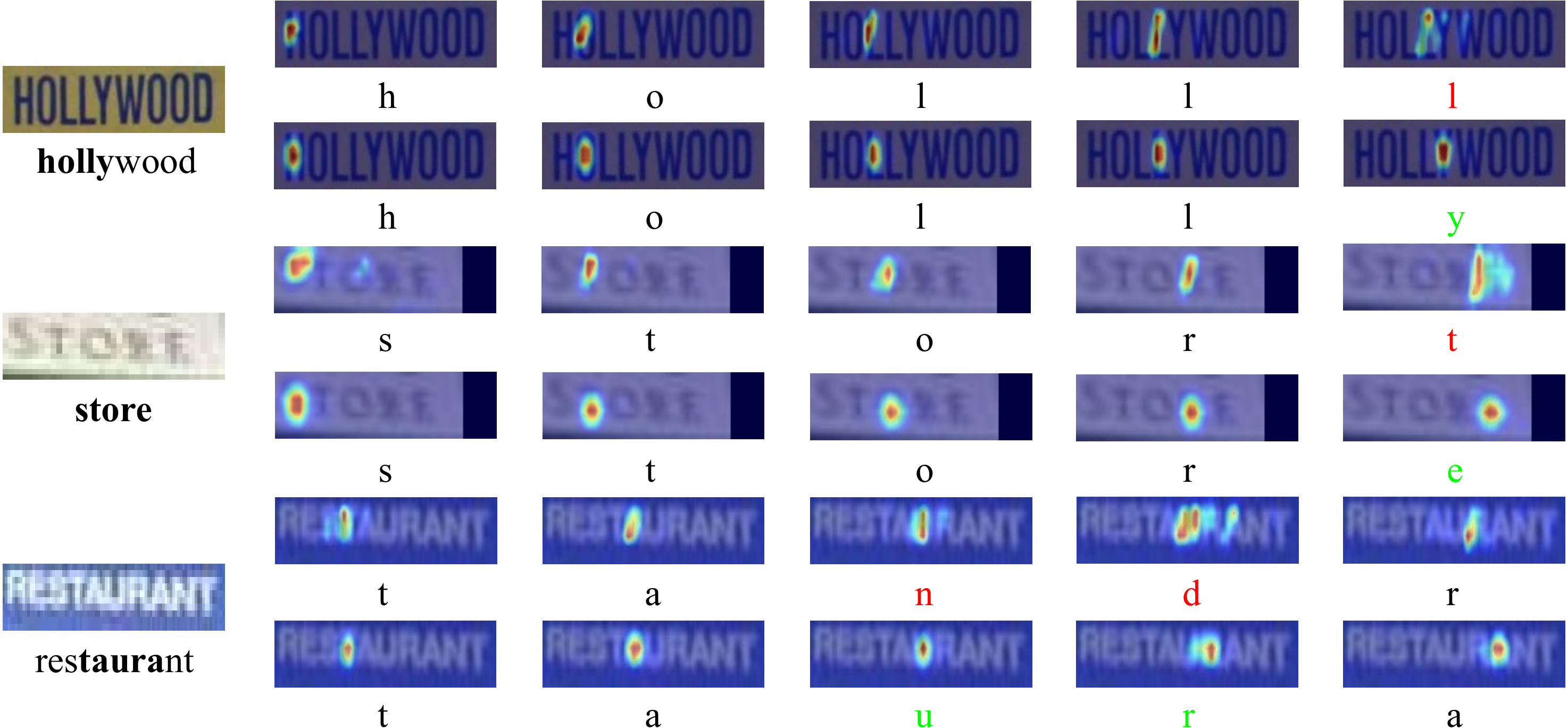}
   \caption{The visualization of the attention maps and corresponding recognition results. For each word sample, the first line is the attention maps of the traditional attention mechanism, and the second line is the attention maps of our proposed GCAN. Due to the space limitation, only the 5 attention maps are presented.}
   \label{fig_att}
\end{figure*}

\subsection{Text Recognition Performance}
\label{performance}

We compare our GCAN with the previous works on public benchmarks. As shown in TABLE~\ref{tabel_sota}, GCAN achieves best performance on two datasets and second best performance on three datasets, which verifies the effectiveness of our methods.

For regular datasets, our proposed GCAN achieves state-of-the-art results on IIIT5k and the second best result on SVT only $0.1\%$ lower compared with state-of-the-art method ESIR~\cite{zhan2019esir}. Compared with the SAR~\cite{li2019show} reproduced by ourselves, which acts as a baseline, our method outperforms it on all three regular datasets, especially on SVT for $3.4\%$ (from $86.7\%$ to $90.1\%$) performance gain.

For irregular datasets, our proposed method also performs better compared with the reproduced SAR on all three benchmarks. For example, our method improves $4.2\%$ (from $77.0\%$ to $81.2\%$) and $1.9\%$ (from $83.7\%$ to $85.6\%$) on SVTP and CUTE respectively. Compared with other methods, we get the state-of-the-art performance on SVTP and second better result on the other two benchmarks, which shows the effectiveness of our proposed method to deal with irregular text.

Some qualitative results are shown in Fig.~\ref{fig_att}, our proposed GCAN predicts more concentrated attention weights and achieves better recognition performance accordingly. 

\subsection{Ablation Study}
\label{ablation}

In order to show the effectiveness of our proposed GCRM, we conduct an ablation study here. As shown in TABLE~\ref{tabel_ablation}, the model with GCRM outperforms the baseline SAR with or without supervision of attention weights. Note that SAR with simple supervision on the attention weights does not improve the performance significantly, only $0.1\%$ (from $93.0\%$ to $93.1\%$) and $0.6\%$ (from $86.7\%$ to $87.3\%$) on IIIT5K and SVT. However, the model with GCRM achieves more performance gain when the attention weights are supervised. We explain that GCRM can be trained better with the supervision and achieves better performance of the refinement. It also shows that the explicit constraint plays an important role for the attention based methods.

\begin{table}[t]
\caption{Performance comparison about the proposed GCRM. Estimation means that estimating the Gaussian parameters instead of predicting. $L_{att}$ represents supervising the attention weights}
\label{tabel_ablation}
\centering
   \begin{tabular}{c|c|c|c|c|c}
   \hline Methods    & $L_{att}$  & IIIT5K        & SVT           & SVTP & IC15 \\
   \hline \hline SAR-reproduced  &            & 93.0          & 86.7          & 77.0 & 76.1 \\
   SAR-reproduced         & \checkmark & 93.1          & 87.3          & 78.8 & 75.4 \\
   
   \hline with Estimation  &            & 93.4          & 88.1          & 78.4 & 75.6 \\
   with Estimation         & \checkmark & 93.8          & 88.4          & 78.8 & \textbf{77.5} \\
   \hline with GCRM         &            & 93.6          & 86.9          & 79.1 & 77.0  \\
   with GCRM         & \checkmark & \textbf{94.4} & \textbf{90.1} & \textbf{81.2} &77.1 \\
   \hline  
   \end{tabular}
\end{table}

\begin{table}[t]
\caption{ Comparison of speed during training and inference}
\label{speed_rebuttal}
\centering
   \begin{tabular}{|c|c|c|}
   \hline Methods & Training & Inference\\
   \hline
   \hline SAR-reproduced & 67.9ms & 45.4ms\\
   GCAN & 75.5ms & 54.3ms\\
   \hline
   \end{tabular}
\end{table}

Another possible solution is to estimate the Gaussian parameters instead of predicting with GCRM. Specifically, we estimate the mean and variance from the raw attention weights with maximum likelihood estimation, then we use them to construct Gaussian mask to refine the raw attention weights. As shown in TABLE~\ref{tabel_ablation} our GCRM still works better in three benchmarks than estimation, which shows the effectiveness of GCRM. One possible explanation is that when facing attention diffusion the estimated parameters are still inaccurate, so the estimated Gaussian masks can not make the raw attention weights more concentrated.


\subsection{Training and Inference Speed}
\label{speed}

The proposed GCRM introduces extra calculation in the attention operation, we conduct experiments to analysis the influence on training and inference speed. We record the speed of two methods on the same hardware (NVIDIA M40), the time consumed by a single image of inference is averaged over six benchmarks. As shown in TABLE~\ref{speed_rebuttal}, GCAN only consumes less than 10ms more during both training and inference. Considering the performance improvement, the extra time consumption is acceptable.

\begin{figure}[h]
\begin{center}
   \includegraphics[width=1.0\linewidth]{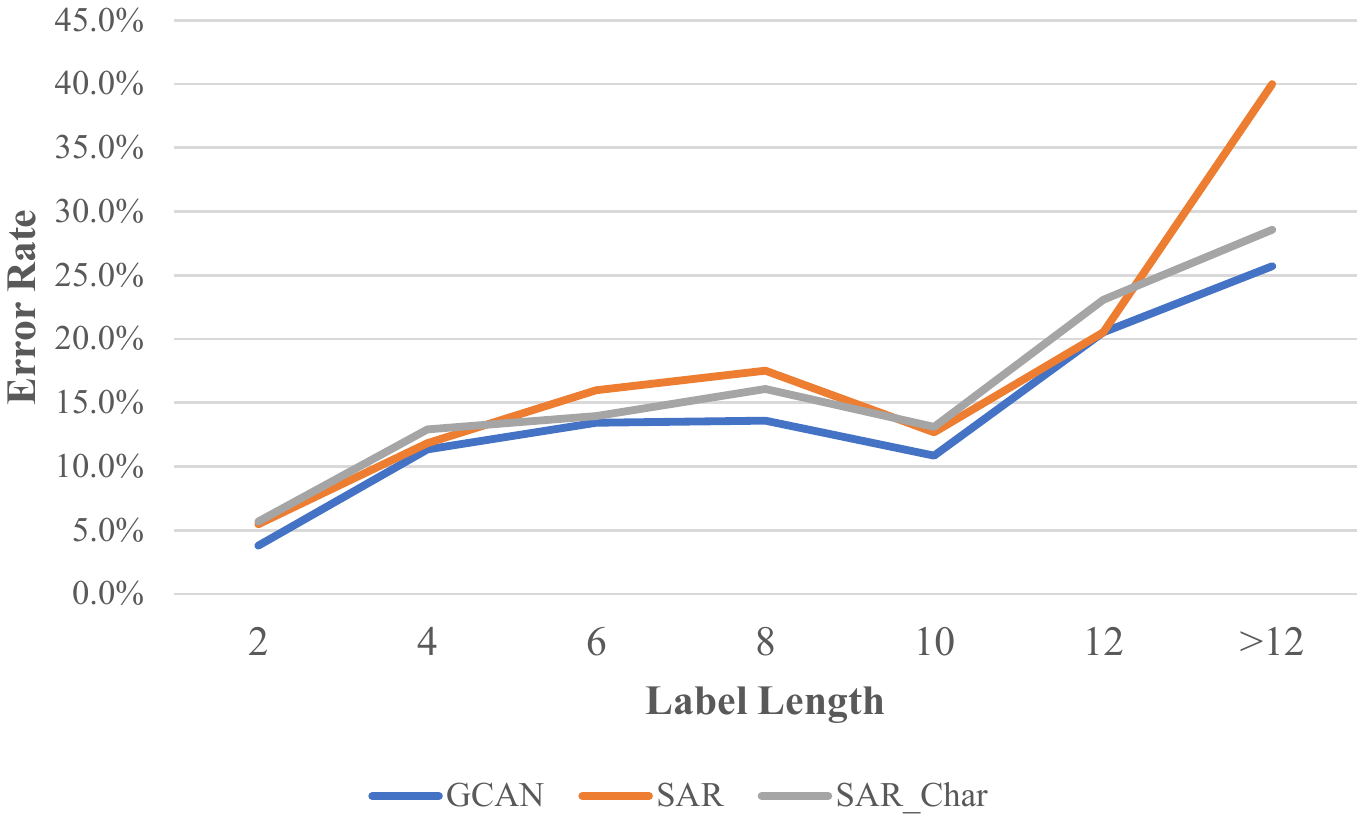}
\end{center}
   \caption{The error rate of the three methods as the label length grows.}
\label{fig_length}
\end{figure}

\subsection{Performance Comparison of Different Text Length}
\label{length}

As we know, attention based methods struggle in the situation of long text. We argue that conventional attention mechanism suffers from attention alignments without explicit constraints. Here an experiment is performed to show the robustness of our GCAN when handling long text. As shown in Fig.~\ref{fig_length}, we analyze error rate according to different label length on six benchmarks, where $SAR\_Char$ indicates SAR with the supervision of attention weights. When the label length is shorter than $6$, the performance of three methods seems similar. However, as the length increases, our GCAN outperforms other two methods significantly. Compared with simple supervision, the methods benefits a lot from the explicit constraints, which also shows the effectiveness of our methods. 


\section{Conclusion}
In this paper, we introduce the problem of \textit{attention diffusion} in the existing attention-based methods. The \textit{attention diffusion} is that the attention maps are not concentratedly distributed and it may introduce noise features into the decoding process. To solve this problem, we propose the Gaussian Constrained Refinement Module to predict an additional Gaussian mask which is applied to refine the raw attention map later. Integrated with the proposed module, our proposed Gaussian Constrained Attention Network achieves convincing performance gain compared with conventional attention mechanism and gets state-of-the-art results on several benchmarks.



\section*{Acknowledgment}

Supported by the National Key R\&D Program of China (2017YFB1002400), the Open Research Project of the State Key Laboratory of Media Convergence and Communication, Communication University of China, China (No. SKLMCC2020KF004), the Beijing Municipal Science \& Technology Commission (Z191100007119002), the CCF-Tencent Open Fund, the Youth Innovation Promotion Association CAS, the Excellent Talent Introduction of Institute of Information Engineering of CAS (No. Y7Z0111107), and the Key Research Program of Frontier Sciences, CAS, Grant NO ZDBS-LY-7024, the National Natural Science Foundation of China (No. 62006221). We thank the anonymous reviewers for their thoughtful comments.






{
  \bibliographystyle{IEEEtran}
  \bibliography{refer.bib}
}
%



\end{document}